\documentclass{article} % For LaTeX2e
\usepackage{iclr2026_conference,times}

\usepackage{amsmath,amsfonts,bm}

\def\eqref#1{equation~\ref{#1}}
\def\1{\bm{1}}

\DeclareMathAlphabet{\mathsfit}{\encodingdefault}{\sfdefault}{m}{sl}
\SetMathAlphabet{\mathsfit}{bold}{\encodingdefault}{\sfdefault}{bx}{n}

\usepackage{hyperref}
\usepackage{url}
\usepackage{graphicx}
\usepackage{amsmath,amssymb}
\usepackage{color}
\usepackage{wrapfig}
\usepackage{epsfig}
\usepackage{multirow}
\usepackage{booktabs}
\usepackage{listings}
\usepackage{algorithm}
\usepackage{algorithmic}
\usepackage{arydshln}
\usepackage{threeparttable}
\usepackage{colortbl}
\usepackage{enumitem}
\usepackage{siunitx}
\usepackage{placeins}
\usepackage{bm}
\usepackage{makecell}
\usepackage{array}

\usepackage{soul}

\usepackage{caption}
\usepackage{dblfloatfix}
\usepackage{pifont}

\usepackage[export]{adjustbox}

\usepackage{subcaption}
\usepackage{tabularx}
\usepackage{tcolorbox}
\usepackage{footnote} 
\let\svthefootnote\thefootnote
\usepackage{xcolor}
\definecolor{myblue}{rgb}{0.97,0.978,0.999}
\definecolor{mygreen}{rgb}{0.961,0.985,0.97}
\definecolor{mygray}{gray}{.9}
\definecolor{ggray}{RGB}{127,127,127}
\definecolor{reda}{RGB}{192,0,0}
\definecolor{redb}{RGB}{217,148,143}
\definecolor{lineblue}{HTML}{3670AF}
\definecolor{backblue}{HTML}{239DE0}
\definecolor{turquoise}{HTML}{BCE6F9}
\definecolor{textblue}{HTML}{316A7B}
\definecolor{linegreen}{HTML}{379E7E}
\definecolor{linepink}{HTML}{DC5A5A}
\definecolor{correct}{HTML}{00B050}
\definecolor{wrong}{HTML}{EE2025}
\definecolor{myyellow}{RGB}{190,144,0}
\definecolor{mygreen}{RGB}{80,100,40}
\definecolor{myblue}{RGB}{30,90,100}
\definecolor{codegreen}{RGB}{79,126,127}
\definecolor{codedefine}{RGB}{153,54,159}
\definecolor{codefunc}{RGB}{73,122,234}
\definecolor{codecall}{RGB}{73,122,234}
\definecolor{codepro}{RGB}{212,96,80}
\definecolor{codedim}{RGB}{89,152,195}
\definecolor{gold}{RGB}{234, 51, 35}

\definecolor{mygreen2}{RGB}{80,100,40}

\usepackage{soul} 
\usepackage{lipsum} 

\definecolor{myhighlight}{RGB}{73, 108, 136}

\newcommand{\ie}{\textit{i}.\textit{e}.}
\newcommand{\cf}{\textit{cf}. }
\newcommand{\eg}{\textit{e}.\textit{g}.}

\newcommand{\method}{\textsc{SinkTrack}}
\makeatletter
\newcommand{\thickhline}{
    \noalign {\ifnum 0=`}\fi \hrule height 1pt
    \futurelet \reserved@a \@xhline
}

\definecolor{mypink}{HTML}{023e8a}

\title{SinkTrack: Attention Sink based Context \\ Anchoring for Large Language Models}

\author{Xu Liu, Guikun Chen, Wenguan Wang$^{\dag}$ \\
    The State Key Lab of Brain-Machine Intelligence, Zhejiang University \\
}
\iclrfinalcopy
\begin{document}

\maketitle

\begin{abstract}
Large language models (LLMs) suffer from hallucination and context forgetting. Prior studies suggest that \textit{attention drift} is a primary cause of these problems, where LLMs' focus shifts towards newly generated tokens and away from the initial input context. To counteract this, we make use of a related, intrinsic characteristic of LLMs: \textit{attention sink} -- the tendency to consistently allocate high attention to the very first token (\ie, $\langle$BOS$\rangle$) of a sequence. Concretely, we propose an advanced context anchoring method, \method{}, which treats $\langle$BOS$\rangle$ as an information anchor and injects key contextual features (such as those derived from the input image or instruction) into its representation. As such, LLM remains anchored to the initial input context throughout the entire generation process. \method{} is \textit{training-free}, \textit{plug-and-play}, and introduces \textit{negligible inference overhead}.  Experiments demonstrate that \method{} mitigates hallucination and context forgetting across both textual (\eg, \textbf{+21.6}\% on SQuAD2.0 with \textit{Llama3.1-8B-Instruct}) and multi-modal (\eg, \textbf{+22.8}\% on M3CoT with \textit{Qwen2.5-VL-7B-Instruct}) tasks. Its consistent gains across different architectures and scales underscore the robustness and generalizability. We also analyze its underlying working mechanism from the perspective of information delivery. Our source code is available at \href{ https://github.com/67L1/SinkTrack}{GitHub}.

\end{abstract}
\renewcommand{\thefootnote}{}
\footnotetext[1]{\dag\hspace{0.2em}Corresponding Author: Wenguan Wang.}
\renewcommand{\thefootnote}{\svthefootnote}

\section{Introduction}
\label{sec:intro}
\begin{wrapfigure}[17]{r}{0.48\textwidth}
    \vspace{-10pt}
    \centering
    \small
    \includegraphics[width=0.97\linewidth]{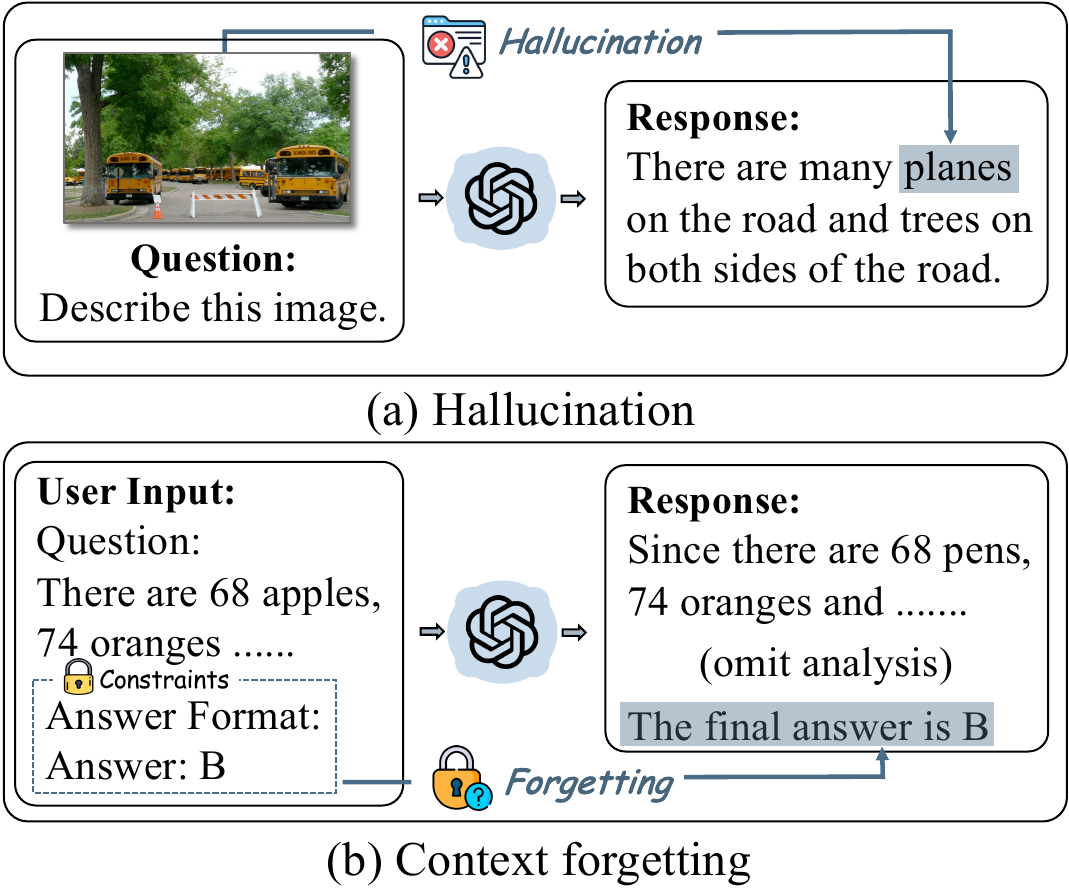}
    \captionsetup{font=small,width=0.45\textwidth}
    \vspace{-6pt}
    \caption{Hallucination and context forgetting are pervasive obstacles in LLMs.}
    \label{fig:intro}
\end{wrapfigure}
\textbf{Background.}
Large language models (LLMs) have demonstrated impressive capabilities in textual and multi-modal understanding ~\citep{bevilacqua2025automated,xu2024llm4workflow,chen2024scene,yang2024doraemongpt,li2023zero,li2024survey,li2025compositional,li2026relation}. However, their deployment in critical applications is constrained by two hard, long-standing problems: \textit{hallucination}~\citep{chang2023surveyevaluationlargelanguage,zhang2025cchallnovelbenchmarkjoint,rawte2023surveyhallucinationlargefoundation} and \textit{context forgetting}~\citep{liu2023lostmiddlelanguagemodels,ling2023domain,li2025singleturnsurveymultiturninteractions}. Hallucination is the generation of content factually inconsistent with the provided input. For instance,  when asked to describe an image of buses on the road, LLM might identify ``planes'' that do not exist in the image (\cf Fig.~\ref{fig:intro}a). Context forgetting occurs during long-form tasks, where LLM may lose track of initial instructions, resulting in unexpected answers. In a multi-step reasoning dialogue, LLM might lose track of the initial instruction on answer formation (\cf Fig.~\ref{fig:intro}b).
% These problems expose core deficiencies in the faithfulness and long-context robustness of LLMs, which constrain their deployment in critical applications~\citep{Huang_2025,liu2025comprehensivesurveylongcontext}.

\begin{wrapfigure}[17]{r}{0.4\textwidth}
    % \vspace{-7pt}
    \centering
    \small
    \includegraphics[width=0.95\linewidth]{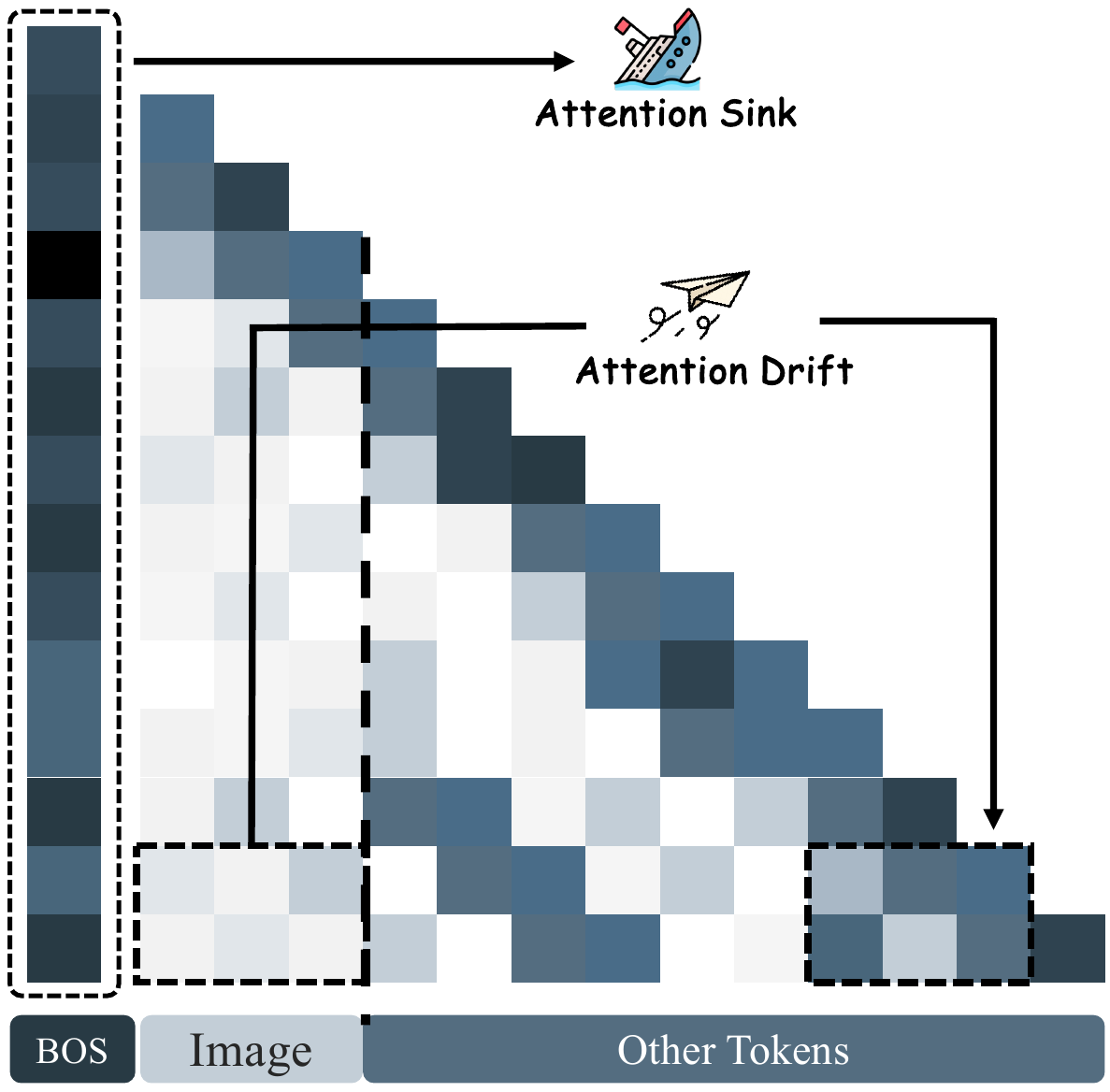}
    \captionsetup{font=small,width=0.36\textwidth}
    \vspace{-7pt}
    \caption{Attention drift and sink (the \textcolor{white}{\hl{darker}} the color, the greater the weight).}

    \label{fig:att_drift}
\end{wrapfigure}

\textbf{Motivation.}
Prior studies~\citep{peysakhovich2023attentionsortingcombatsrecency,an2025mitigatingobjecthallucinationslarge,liu2023lostmiddlelanguagemodels} identify \textit{attention drift} as a direct cause of hallucination and context forgetting (\cf Fig.~\ref{fig:att_drift}). As LLM generates a sequence, its attention progressively shifts towards newly created tokens, while attention on the initial tokens -- which often contain the most critical input information, such as image features or user instructions -- dramatically decays. Interestingly, we observe that this drift coexists with a countervailing phenomenon: the \textit{attention sink}~\citep{xiao2024efficientstreaminglanguagemodels,a2}. While attention on most initial tokens fades, the very first token (\ie, $\langle$BOS$\rangle$) intrinsically main-
\sethlcolor{myhighlight}
tains a high degree of attention throughout generation, even though it is semantically sparse (\cf $1_{st}$ column in Fig.~\ref{fig:att_drift}). These two patterns, one of decay and one of stability, present a synergistic opportunity. The stability of the attention sink provides a natural mechanism to counteract the information loss caused by attention drift. Based on this insight, our core idea is to transform the passive attention sink into an active mechanism for context anchoring (\ie, using $\langle$BOS$\rangle$ as an information anchor), thereby mitigating the information decay caused by attention drift. Through injecting key contextual features (such as those derived from the input image or instruction) into the representation of $\langle$BOS$\rangle$, LLM remains anchored to the initial input context throughout generation.

% Rather than counteract this drift~\citep{NEURIPS2024_91315fbb,yu2025mitigatepositionbiaslarge}, we make use of a related, intrinsic characteristic of LLMs: \textit{attention sink}~\citep{xiao2024efficientstreaminglanguagemodels,a2} -- the tendency to consistently allocate high attention to the very first token (\ie, $\langle$BOS$\rangle$) of a sequence. Despite the overall drift, $\langle$BOS$\rangle$ consistently retains high attention, yet it carries almost no semantic information, creating a ``high-weight, low-information'' anomaly (\cf Fig.~\ref{fig:att_drift}).

% Our core idea is to transform the passive attention sink into an active mechanism for context anchoring (\ie, using $\langle$BOS$\rangle$ as an information anchor), thereby mitigating the information decay caused by attention drift. Through injecting key contextual features (such as those derived from the input image or instruction) into the representation of $\langle$BOS$\rangle$, LLM remains anchored to the initial input context throughout the entire generation process.

\textbf{Methodology.}
To implement the above idea, \ie, using $\langle$BOS$\rangle$ as an information anchor, we make an iterative exploration. The exploration begins with a \texttt{hard injection} (\S\ref{sec:hard}), which directly replaces the Value vector of $\langle$BOS$\rangle$ in the KV cache with a mean-pooled information vector (\eg, from image or instruction features). This implementation leads to a collapse, revealing that LLM's original computational flow should be preserved. We then investigate a \texttt{soft injection} (\S\ref{sec:soft}), which blends the pooled vector into LLMs using a weighted sum to respect the original flow. Despite effective, there are concerns regarding its robustness: it relies on manually tuned hyperparameters and injects all pooled representations into $\langle$BOS$\rangle$ \textit{equally} -- which might add noise and hurt performance on long contexts. Motivated by these limitations, our final solution is \method{} (\S\ref{sec:mean}), which is built upon a dual-track cross-attention mechanism. In this design, one track enables $\langle$BOS$\rangle$ to adaptively query and retrieve relevant information from the context (\ie, cross-attention), while a parallel self-attention track preserves LLM's native computational flow, leading to consistent performance gains.

% Such design allows adaptive injection, but performance still dropped as context length increased. We found this was due to mean-pooling the source context, which loses detail. In our final version (\S\ref{sec:womean}), we remove mean-pooling entirely. Instead, we use a key property of cross-attention: the query and context do not need to be the same length. This lets the first token attend directly to the full, raw source context, with no compression and no information loss.

\textbf{Merits.}
Through our study of anchor formulation, injection strategies, and source integrity, we design \method{} as a lightweight, adaptive enhancement with three benefits: First, it is \textit{plug-and-play}. As a training-free, inference-time enhancement, \method{} can be applied to any pre-trained LLM without the risks of fine-tuning, \eg, catastrophic forgetting and model collapse. Second, it is \textit{efficient}. Its core intervention is a one-off, lightweight operation performed prior to generation. Since the intervention can be stored in KV cache, \method{} adds negligible computational overhead to the subsequent auto-regressive steps, thereby preserving LLMs' original inference speed. Third, it is \textit{generalizable}. \method{} makes use of \textit{attention sink}, a common, intrinsic characteristic of decoder-only LLMs and multi-modal LLMs (MLLMs), to mitigate hallucination and context forgetting. This principle design makes it inherently generalizable across architectures, scales, and modalities, as validated in experiments (\S\ref{sec:exp}) on different LLMs and MLLMs from 3B to 12B sizes.

\textbf{Results and Analysis.}
\method{} demonstrates consistent performance improvements over both direct prompting and chain-of-thought (\texttt{CoT}) reasoning across various multi-modal and text-based benchmarks (\S\ref{sec:exp_quan_res}). For instance, on the multi-modal M3CoT benchmark~\citep{m3cot}, \method{} increases the accuracy of \textit{Qwen2.5-VL-7B} by \textbf{22.8}\% over \texttt{CoT}, and on the reading comprehension benchmark, SQuAD2.0~\citep{squad2}, it improves \textit{Llama3.1-8B}'s accuracy by \textbf{21.6}\%. Qualitative analysis reveals that \method{} enhances the logical structure and detail focus in LLM's reasoning process (\S\ref{sec:exp_quali_res}). We also explore the underlying working mechanism of \method{} (\S\ref{sec:analysis}).
\vspace{-2pt}
\section{Methodology}
\vspace{-2pt}
\label{sec:metho}

\begin{figure}[t]
    \vspace{-24pt}
    \centering
    % \vspace{-4pt}
    \includegraphics[width=0.97\linewidth]{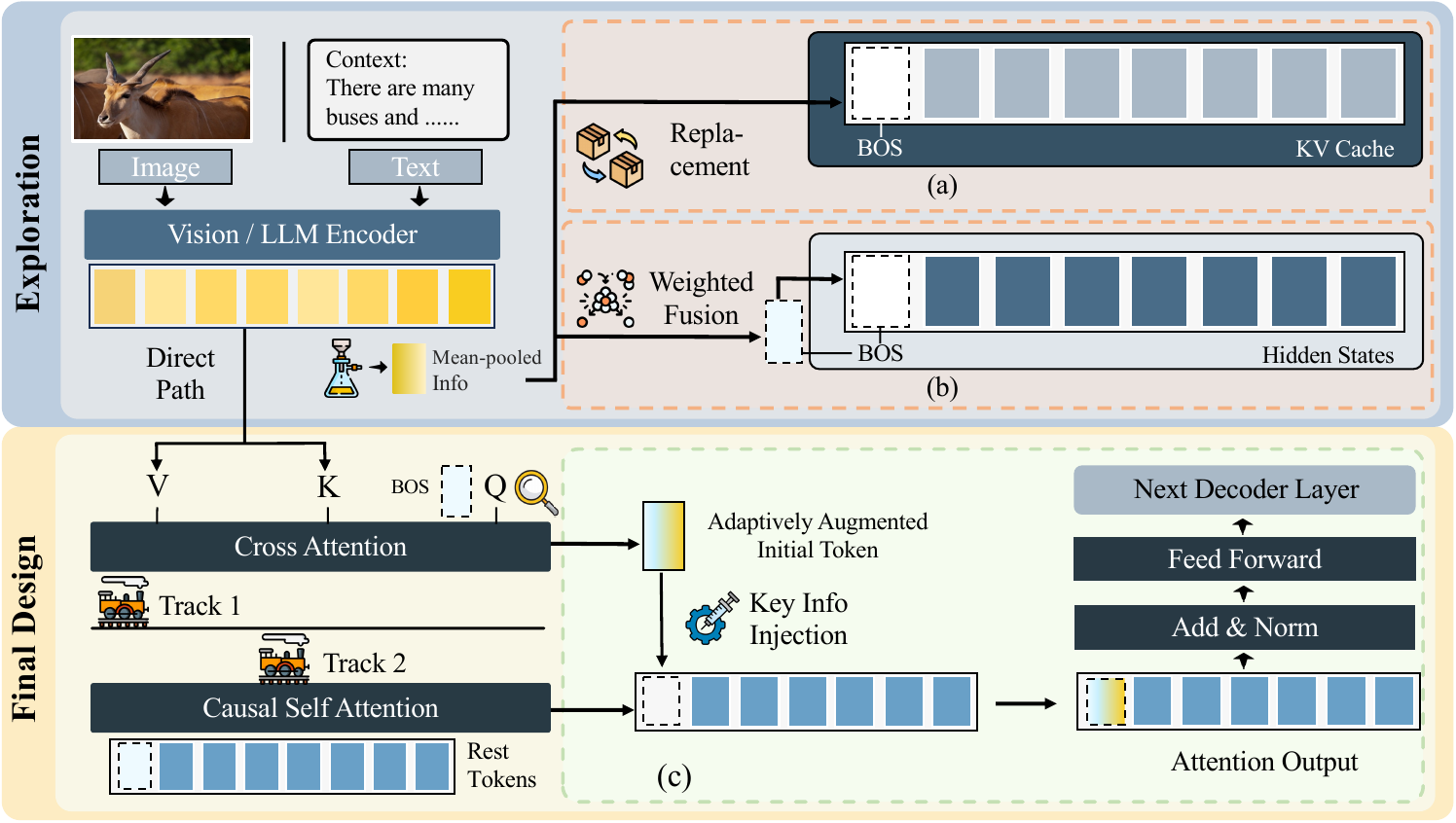}
    \put(-130,165.4){\footnotesize \texttt{Hard Injection} (\S\ref{sec:hard})}
    \put(-130,112.2){\footnotesize \texttt{Soft Injection} (\S\ref{sec:soft})}
    \put(-200.5,8.8){\footnotesize \method{} (\S\ref{sec:mean})}
    
    \vspace{-4pt}
    \caption{An overview of our methodology exploration.}
    \label{fig:overview}
    \vspace{-8pt}
\end{figure}

% This section details the proposed \method{}. 
We first review the phenomena of \textit{attention drift} and \textit{attention sink} that motivate our method (\S\ref{sec:twoatt}). Then, we present the exploration undertaken to use \textit{attention sink} as a dynamic information anchor, beginning with an initial \texttt{hard injection} (\S\ref{sec:hard}), progressing to a \texttt{soft injection} (\S\ref{sec:soft}), and leading to the development of \method{}, our core \textit{dual-track} mechanism (\S\ref{sec:mean}). 
% This path culminates in the final design, which \textbf{\textit{eliminates a lossy pooling operation}}(\S\ref{sec:womean}).

\vspace{-20pt}
\subsection{Attention Patterns of LLMs}
\vspace{-2pt}
\label{sec:twoatt}

During auto-regressive generation, LLMs exhibit two fundamental and co-existing attention patterns that motivate our approach. The first is \textit{attention drift}~\citep{peysakhovich2023attentionsortingcombatsrecency,an2025mitigatingobjecthallucinationslarge,liu2023lostmiddlelanguagemodels} (\cf Fig.~\ref{fig:att_drift}), where LLMs' attention progressively concentrates on more recent tokens. As the generation proceeds, attention on the initial tokens, which contain the primary context (\eg, instructions or image), dramatically decays. This progressive loss of focus on the original input is a primary cause of hallucination and context forgetting. The second pattern is \textit{attention sink}~\citep{xiao2024efficientstreaminglanguagemodels,a2}. (\cf $1_{st}$ column in Fig.~\ref{fig:att_drift}). In direct contrast to the general decay of attention on initial tokens, the very first token of the sequence (\ie, $\langle$BOS$\rangle$) intrinsically and consistently maintains an exceptionally high degree of attention throughout the generation process. $\langle$BOS$\rangle$ is semantically sparse, yet it functions as LLMs’ most stable point of attention.

These two patterns, one of decay and one of stability, operate in tandem. Surprisingly, while these phenomena have been studied independently, their synergistic potential has been overlooked. This presents a clear opportunity: rather than attempting to alter these inherent behaviors, this work, to our best knowledge, is the first to leverage their interplay. Our core idea is to repurpose the passive attention sink into an active mechanism to counteract drift. Detailed implementations are given next.

\subsection{Hard Injection via Direct Replacement}
\label{sec:hard}
% \begin{wrapfigure}[8]{r}{0.45\textwidth}
%     \vspace{-12pt}
%     \centering
%     \small
%     \includegraphics[width=\linewidth]{figs/hard.pdf}
%     \captionsetup{font=small,width=0.43\textwidth}
%     \vspace{-17pt}
%     \caption{Hard injection strategy causes model collapse.}

%     \label{fig:hard}
% \end{wrapfigure}

This work targets \textbf{decoder-only LLM}, which is composed of $L$ Transformer decoder blocks stacked sequentially. The $l$-th block processes the sequence of hidden states from the preceding block, $\bm{H}^{(l-1)} \in \mathbb{R}^{n \times d}$, where $n$ is the sequence length and $d$ is the hidden dimension. Each decoder block contains two main sub-layers. The first is a multi-head attention (MHA) sub-layer: 
\begin{equation}
\bm{M}^{(l)} = \text{LayerNorm}\big(\bm{H}^{(l-1)} + \text{MHA}(Q = \bm{H}^{(l-1)},\; K = \bm{H}^{(l-1)},\; V = \bm{H}^{(l-1)})\big).
\label{eq:mha_with_addnorm}
\end{equation}
The second sub-layer is a feed-forward network (FFN):
\begin{equation}
\bm{H}^{(l)} = \text{LayerNorm}(\bm{M}^{(l)} + \text{FFN}(\bm{M}^{(l)})).
\label{eq:ffn_with_addnorm}
\end{equation}
\textbf{Implementation.}
Based on the characteristic of attention sink, the most direct and simple method to implement our idea is to place information at the representation of $\langle$BOS$\rangle$. Therefore, we first attempt \texttt{hard injection} (\cf Fig.~\ref{fig:overview}a), which directly manipulates the embedding of $\langle$BOS$\rangle$. Concretely, we directly replace the Value vector ($V$ in Eq.~\ref{eq:mha_with_addnorm}) corresponding to $\langle$BOS$\rangle$ in the KV cache with an external information vector $\bm{f}_{\text{info}}$. $\bm{f}_{\text{info}}$ is derived from the input context. For the textual task, it is the prompt embeddings; for the multi-modal task, it is the visual features from the MLLM's own vision encoder. It is then compressed via mean pooling to match the dimensionality.

% This injected vector, derived from either textual or image features, is compressed via mean pooling to match the required dimensionality.

% As shown in , within the model's attention module, we directly replace the Value (V) vector corresponding to the initial token in the KV cache with an external key information vector. To perform this replacement, the key information is compressed into a single vector via mean pooling to match the dimensionality of the initial token's vector.

\noindent\textbf{Results and Analysis.} Unfortunately, \texttt{hard injection} leads to a complete model collapse, yielding meaningless outputs. This indicates that LLM's internal computational pathways and learned attention patterns are sensitive to direct manipulation. Abruptly overwriting values in the KV cache disrupts this finely-tuned structure, leading to a severe degradation. This finding establishes a core principle: any effective information injection must preserve the integrity of the LLM's native computational flow, motivating our development of the more nuanced \texttt{soft injection}.

% This failed experiment reveals a critical lesson: the internal computational flow and attention patterns learned by LLMs during training are extremely fine-grained and fragile. Any operation that crudely disrupts this inherent flow can lead to detrimental consequences. The failure inspires that any effective information injection must preserve the model's pretrained sequence structure, which guides us towards a more moderate \texttt{soft injection} approach.

\vspace{-5pt}
\subsection{Soft Injection via Static Fusion}
\label{sec:soft}
\vspace{-3pt}

\noindent\textbf{Implementation.}
In contrast to the disruptive replacement used in \texttt{hard in  jection}, \texttt{soft injection} fuses information into LLMs' computational flow. This is achieved by fusing a mean-pooled information vector with the original hidden state of $\langle$BOS$\rangle$ via weighted sum (\cf Fig.~\ref{fig:overview}b):
% \begin{equation}
% \bm{H}'_{0} = \alpha \cdot \bm{H}_{0} + (1 - \alpha) \cdot \bm{f}_{\text{info}},
% \label{eq:soft_injection}
% \end{equation}
\begin{equation}
\bm{\bar{H}}^{(l)}_{0} = \alpha \cdot \bm{H}^{(l)}_{0} + (1 - \alpha) \cdot \bm{f}_{\text{info}},
\label{eq:soft_injection}
\end{equation}
where $\bm{H}^{(l)}_{0} $ is the original hidden state of the first token $\langle$BOS$\rangle$ at injection layer $l$, which corresponds to the output of the FFN sub-layer (Eq.~\ref{eq:ffn_with_addnorm}). $\bm{f}_{\text{info}} \!\in\! \mathbb{R}^{d}$ is the mean-pooled information vector, $\alpha$ is the injection strength, and $\bm{\bar{H}}^{(l)}_{0}$ is the updated hidden state after \texttt{soft injection}. Such an injection strategy holds two main hyperparameters: the injection layers and the injection strength $\alpha$.

% \setlength{\columnsep}{10pt}
% \begin{wrapfigure}{r}{0.60\linewidth}
%     \vspace{-48pt}
%     \centering
%     \includegraphics[width=\linewidth]{figs/softall.pdf}
%     \vspace{-49pt}
%     \caption{Soft injection performance.}
%     \label{fig:soft}
%     \vspace{-20pt}
% \end{wrapfigure}

% \begin{wrapfigure}[12]{r}{0.7\textwidth}
%     \vspace{-9pt}
%     \centering
%     \small
%     \includegraphics[width=\linewidth]{figs/softall.pdf}
%     \captionsetup{font=small,width=0.4\textwidth}
%     \vspace{-17pt}
%     \caption{Performance of Soft injection.}
%     \label{fig:soft}
% \end{wrapfigure}

\begin{figure}[t]
    \centering
    \vspace{-21pt}
    \includegraphics[width=0.97\linewidth]{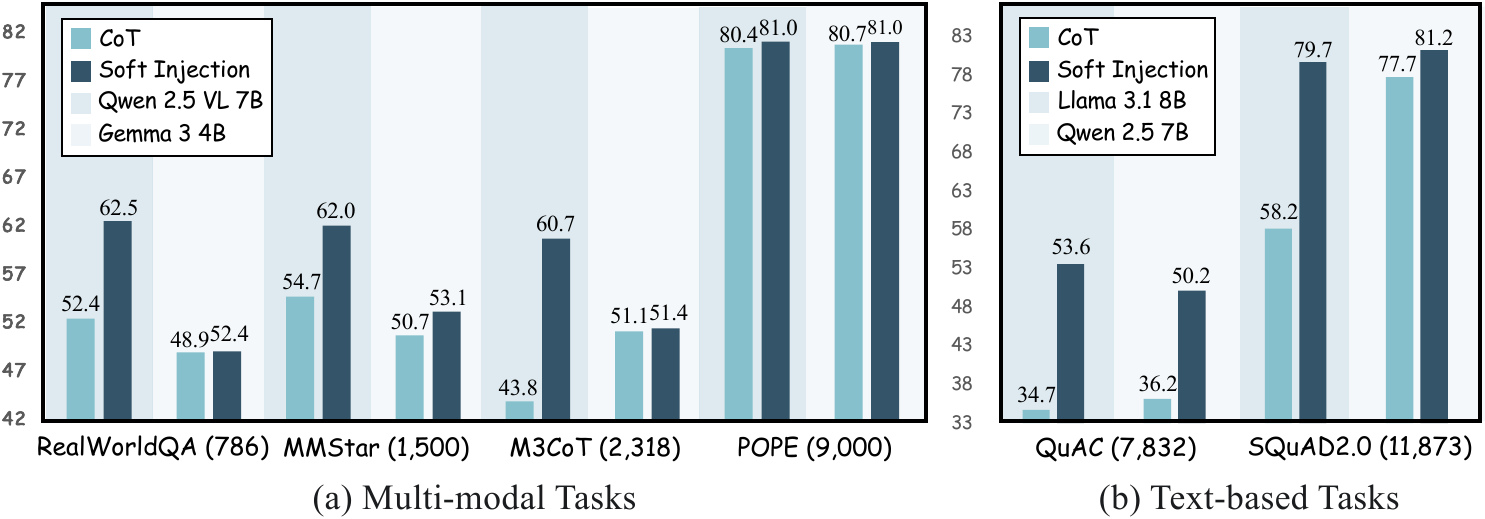}
    \vspace{-6pt}
    \caption{\texttt{Soft injection} achieves considerable performance improvements (\S\ref{sec:soft}).}
    \label{fig:soft}
    \vspace{-15pt}
\end{figure}

\noindent\textbf{Results and Analysis.}
We evaluate \texttt{soft injection} in four multi-modal and two text-based tasks (\S\ref{sec:setting}). While soft injection achieves considerable gains (\cf Fig.~\ref{fig:soft}) and validates our core premise, its static design reveals fundamental limitations. Its performance is contingent on a manually tuned hyperparameter (\ie, injection strength $\alpha$), restricting its applicability. Furthermore, by uniformly aggregating all contextual information, it lacks the capacity to adaptively select the most salient content. This indiscriminate fusion can introduce noise and even degrade performance, highlighting the need for a dynamic mechanism that can adaptively query the source context.

\subsection{SinkTrack: Dual-Track Cross-Attention Mechanism}
\label{sec:mean}
% \setlength{\columnsep}{5pt}
% \begin{wrapfigure}{r}{0.55\linewidth}
%     \vspace{-30pt}
%     \centering
%     \includegraphics[width=\linewidth]{figs/A.pdf}
%     \vspace{-30pt}
%     \caption{A performance.}
%     \label{fig:A}
%     \vspace{-17pt}
% \end{wrapfigure}
\begin{wrapfigure}[14]{r}{0.66\textwidth}
    \vspace{-12pt}
    \centering
    \small
    \includegraphics[width=\linewidth]{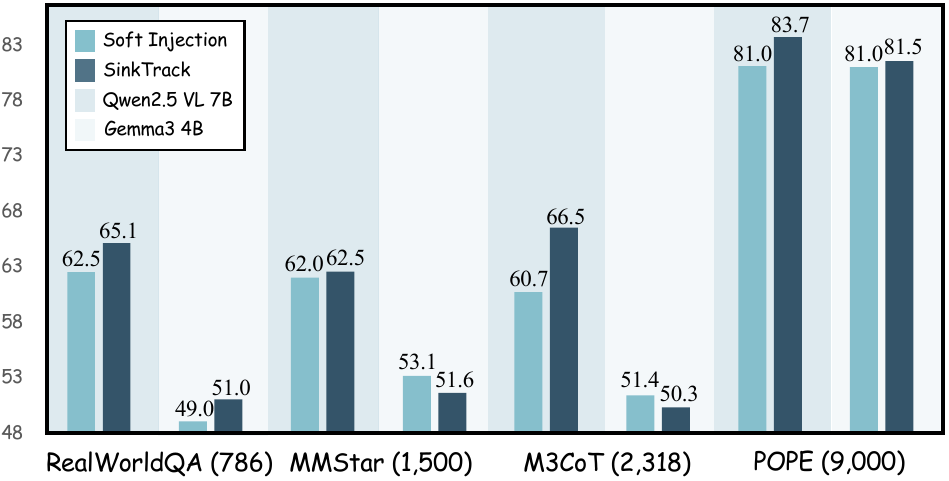}
    \captionsetup{font=small,width=0.5\textwidth}
    \vspace{-17pt}
    \caption{\method{} further improves performance.}
    \label{fig:A}
\end{wrapfigure}

To overcome the limitations of soft injection, we introduce \method{}, \!an \!adaptive \!mechanism centered on a dual-track attention design. \method{} achieves dynamic information infusion by separating the attention computation for $\langle$BOS$\rangle$ from that of all subsequent tokens (\cf \!Fig.~\ref{fig:overview}c). This separation allows \method{} to inject context adaptively while preserving LLM's native computational flow.

% To overcome the limitations of static injection and achieve dynamic, adaptive information infusion, we design \method{}, which is centered on a dual-track cross-attention mechanism. This mechanism divides the model's internal attention flow into two parallel tracks, enabling dynamic information injection while preserving the stability of the model's native structure. The architecture is depicted in Fig.~\ref{fig:overview}c.

\noindent\textbf{Track 1: Adaptive Injection via Cross-Attention}.
The first track performs adaptive injection exclusively for $\langle$BOS$\rangle$. This is achieved by modifying MHA (Eq.~\ref{eq:mha_with_addnorm}) at designated injection layers. Concretely, the hidden state of $\langle$BOS$\rangle$ at injection layer $l$, $\bm{H}^{(l)}_{0}$, serves as the Query ($Q$), while the Key ($K$) and Value ($V$) are now derived from the external, mean-pooled information vector ($\bm{f}_{\text{info}}$). This is implemented by cross-attention:
\begin{equation}
\bm{\bar{H}}^{(l)}_{0} = \text{MHA}(Q = \bm{H}^{(l)}_{0},\; K = \bm{f}_{\text{info}},\; V = \bm{f}_{\text{info}}),
\label{eq:attention_sink}
\end{equation}
where $\bm{\bar{H}}^{(l)}_{0}$ denotes the updated hidden state after cross-attention. While the cross-attention is designed for adaptive injection, one may wonder: how can the low-information query vector $\bm{H}^{(l)}_{0}$ retrieve information adaptively? We emphasize that this adaptive capability is a layer-wise information accumulation. At the initial injection layer, $\bm{H}^{(l)}_{0}$ aggregates a holistic representation of the global context from $\bm{f}_{\text{info}}$. This aggregation infuses $\bm{H}^{(l)}_{0}$ with context, thereby enabling it to formulate targeted queries in subsequent injection layers. This progression allows $\langle$BOS$\rangle$ to identify salient information dynamically, thus obviating the need for a manually tuned injection strength.

% The adaptive nature emerges from an iterative refinement process. At the initial injection layer, the low-information $\bm{H}_{0}$ cannot formulate a specific query. Consequently, it aggregates a broad representation of the global context from $\bm{f}_{\text{info}}$ into itself. In subsequent injection layers, $\bm{H}_{0}$ can then formulate targeted queries to retrieve information. This progression allows the model to identify salient information dynamically, thus obviating the need for a manually tuned injection strength.

% This formulation allows $\langle$BOS$\rangle$ to dynamically query and retrieve the most salient information from the source context, eliminating the need for a manually tuned injection strength.

\begin{figure}[t]
    \centering
    \vspace{-20pt}
    \includegraphics[width=0.97\linewidth]{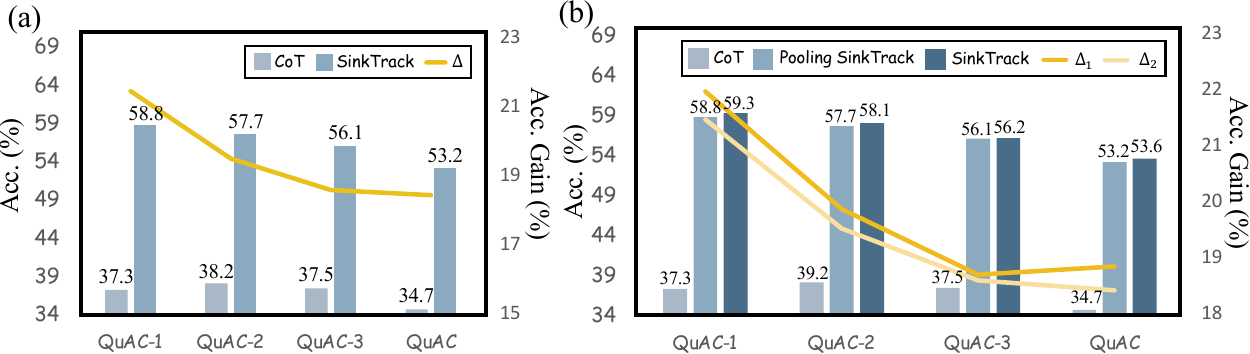}
    \caption{Effect of mean-pooling in information injection (Eq.~\ref{eq:attention_sink}). (a) Using the mean-pooled information vector; (b) using the original (unpooled) one.}
    \label{fig:finalall}
    \vspace{-10pt}
\end{figure}

% \begin{wrapfigure}[14]{r}{0.40\textwidth}
%     \vspace{-12pt}
%     \centering
%     \small
%     \includegraphics[width=\linewidth]{figs/Atext.pdf}
%     \captionsetup{font=small,width=0.4\textwidth}
%     \vspace{-17pt}
%     \caption{Performance of \method{} against \textit{CoT} on QuAC with \texttt{Llama-3.1-8B-Instruct}. The $\Delta$ quantifies the performance improvement of \method{} over the baseline.}
%     \label{fig:Atext}
% \end{wrapfigure}
% \setlength{\columnsep}{-8pt}
% \begin{wrapfigure}{r}{0.45\linewidth}
%     \vspace{-16pt}
%     \centering
%     \includegraphics[width=\linewidth]{figs/Atext.pdf}
%     % \vspace{-30pt}
%     \caption{A performance.}
%     \label{fig:Atext}
%     \vspace{-21pt}
% \end{wrapfigure}

\noindent\textbf{Track 2: Preserving Native Computation via Self-Attention.}
The second track ensures that LLM's computational flow remains undisturbed. All other tokens (\ie, regular tokens) perform the standard causal self-attention computation, attending to the original, unmodified input sequence.
 
% This track's objective is to maintain the linguistic and sequential knowledge learned by the model during pretraining. All other tokens (excluding the initial token) perform a standard causal self-attention computation with the original, unmodified input sequence, including the original initial token. By strictly confining the information injection process to the first track, we ensure that the model's native computational flow remains undisturbed, thereby avoiding the structural disruption caused by hard injection.

\textbf{Sequence Reassembly.}
The two tracks operate in parallel within the same attention block. The final output is a concatenation of the updated $\langle$BOS$\rangle$ representation from Track 1 and the regular token representations from Track 2. This dual-track mechanism allows \method{} to function as an adaptive context anchor, retrieving necessary information without compromising the foundational structure of LLMs. The superiority of this design is demonstrated by its consistent performance gains on long-context benchmarks like QuAC (\cf Fig.~\ref{fig:A}).

\textbf{The Effect of Mean-Pooling.}
Despite outperforming \texttt{CoT}, we observe that \method{}'s performance gains on the long-dialogue QuAC dataset diminish as the context length grows (\cf Fig.\ref{fig:finalall}a). We identify the cause as the mean-pooling operation, a step inherited from \texttt{hard} and \texttt{soft injection}, which inevitably causes information loss by compressing a variable-length sequence into a fixed-size vector -- a problem that intensifies with longer contexts. However, the cross-attention mechanism in \method{} offers a structural advantage, as it does not require sequence length alignment between the Query and the Key/Value. We therefore eliminate the lossy pooling step, instead using the complete, uncompressed sequence of contextual features as the source. This modification effectively mitigates the trend of diminishing gains and preserves performance on long-context tasks (\cf Fig.\ref{fig:finalall}b). The complete process is described in Alg.~\ref{alg:hybrid_attention} to ease understanding.

\section{Experiment}
\label{sec:exp}

% To comprehensively evaluate the effectiveness of the proposed \method{} in mitigating model hallucination and context forgetting, we designed a series of experiments covering different tasks (textual and multi-modal) and LLMs (model sizes ranging from 3B to 12B parameters).
% This section provides a detailed description of the models, baselines, datasets, and implementation details used in our evaluation.

\subsection{Experimental Setup}
\label{sec:setting}

\noindent\textbf{Datasets.} We select four multi-modal datasets and two text-based datasets to assess the performance of \method{} for different modalities.

$\bullet$ \textbf{Multi-modal Datasets.}
Four multi-modal datasets are used for evaluation: \textbf{RealWorldQA}~\citep{realworldqa}, a visual question answering (QA) dataset designed to assess model robustness in complex, real-world scenarios; \textbf{MMStar}~\citep{mmstar}, a comprehensive benchmark for assessing multi-modal capabilities that encompasses 6 core abilities across 18 main tasks; \textbf{M3CoT}~\citep{m3cot}, a task collection focused on multi-modal, multi-hop, complex chain-of-thought reasoning; and \textbf{POPE}~\citep{pope}, a benchmark specifically designed to quantify object-level hallucination in multi-modal models.

$\bullet$ \textbf{Text-based Datasets.}
Two text-based datasets are used for evaluation: \textbf{QuAC}~\citep{quac}, a conversational QA benchmark, to measure the model's ability to handle long-context dependencies. Its structure, where questions build upon the dialogue history, allows us to analyze performance across progressive dialogue turns (QuAC-1, -2, -3) as the context lengthens, thereby assessing the mitigation of context forgetting; and \textbf{SQuAD2.0}~\citep{squad2}, another QA benchmark to test LLMs' ability to abstain from answering when information is absent in the context, providing a direct measure of its capacity to avoid generating fabricated content.

\noindent\textbf{Base LLMs.}
We validate the effectiveness and generalizability of \method{} across mainstream, open-source LLMs and MLLMs. Specifically, \textit{Qwen2.5}~\citep{qwen25}, \textit{MiniCPM3}~\citep{minicpm}, and \textit{Llama3.1}~\citep{llama31} are used for text-based tasks, while \textit{Qwen2.5-VL}~\citep{qwen25vl} and \textit{Gemma3}~\citep{gemma3} are used for multi-modal tasks.

\noindent\textbf{Competitors.}
Since \method{} is training-free and plug-and-play, \texttt{Direct} queries with LLMs, and Chain-of-Thought~\citep{cot} (\texttt{CoT}) are used as our primary competitors. Specifically, CoT guides the model to perform step-by-step reasoning by including instructions, ``Let's think step by step'', in the prompt, without applying any additional information injection mechanism. This comparison allows for a clear isolation of the performance gains attributable to \method{}. 
% Completed experimental results are given in \S\ref{sec:exp_comprehensive}.

\noindent\textbf{Implementation Details.}
We adopt an intermittent injection strategy for both \texttt{soft injection} and \method{}. Specifically, the injection operation is applied at intervals of every 5 layers. This configuration is empirically determined based on our analysis (\S\ref{sec:ablation}), where we observe that such intermittent injection yields better performance compared to dense injection.

\begin{table*}[t]
\vspace{-20pt}
\setlength{\tabcolsep}{7pt}
\centering
\renewcommand{\arraystretch}{1.2}
\begin{adjustbox}{width=0.99\textwidth}
\begin{tabular}{lcccccccccc}
\toprule
\multirow{2}{*}{Model}  &   \multicolumn{2}{c}{\textbf{RealWorldQA (768)}} & \multicolumn{2}{c}{\textbf{MMStar (1,500)}} & \multicolumn{2}{c}{\textbf{M3CoT (2,318)}} & \multicolumn{2}{c}{\textbf{POPE (9,000)}} & \multicolumn{2}{c}{Average} \\
	\cmidrule{2-11}
           &  Acc  &  Macro-F1   & Acc & Macro-F1  &   Acc &  Macro-F1   &  Acc & Macro-F1  &  Acc & Macro-F1  \\
        \midrule
        \rowcolor[rgb]{ .980,  .980,  .980} 
        \multicolumn{11}{c}{\rule{0pt}{2ex}\textit{\fontsize{11pt}{11pt}\selectfont Qwen2.5-VL-3B-Instruct~\citep{qwen25vl}}}     \\  
		\midrule
		\texttt{Direct}  & 31.15\tiny${\textcolor{gray}{\pm1.63}}$ & 23.25\tiny${\textcolor{gray}{\pm4.36}}$  & 24.89\tiny${\textcolor{gray}{\pm1.65}}$ & 39.54\tiny${\textcolor{gray}{\pm1.90}}$  & 33.43\tiny${\textcolor{gray}{\pm0.54}}$ & 44.20\tiny${\textcolor{gray}{\pm3.08}}$  & 53.23\tiny${\textcolor{gray}{\pm0.24}}$ & 69.48\tiny${\textcolor{gray}{\pm0.20}}$  & 35.68 & 44.12\\
		\texttt{CoT}  & 35.69\tiny${\textcolor{gray}{\pm1.61}}$ & 26.46\tiny${\textcolor{gray}{\pm4.71}}$ & 35.24\tiny${\textcolor{gray}{\pm0.83}}$ & 50.68\tiny${\textcolor{gray}{\pm0.95}}$  & 20.69\tiny${\textcolor{gray}{\pm0.51}}$ & 32.04\tiny${\textcolor{gray}{\pm0.62}}$  & 64.58\tiny${\textcolor{gray}{\pm0.22}}$ & 78.10\tiny${\textcolor{gray}{\pm0.16}}$  & 39.05 & 46.82\\
		\texttt{\method{}}   & \textbf{48.28}\tiny${\textcolor{gray}{\pm0.40}}$ & \textbf{28.42}\tiny${\textcolor{gray}{\pm0.22}}$ & \textbf{47.24}\tiny${\textcolor{gray}{\pm0.40}}$ & \textbf{61.38}\tiny${\textcolor{gray}{\pm0.44}}$  & \textbf{48.25}\tiny${\textcolor{gray}{\pm0.09}}$ & \textbf{58.73}\tiny${\textcolor{gray}{\pm1.14}}$  &    \textbf{77.69}\tiny${\textcolor{gray}{\pm0.12}}$ &    \textbf{87.44}\tiny${\textcolor{gray}{\pm0.07}}$  & \textbf{55.37} & \textbf{58.99}\\

        \midrule
        \rowcolor[rgb]{ .980,  .980,  .980} 
        \multicolumn{11}{c}{\rule{0pt}{2ex}\textit{\fontsize{11pt}{11pt}\selectfont Gemma3-4B-Instruct~\citep{gemma3}}}     \\  
		\midrule
		\texttt{Direct}  & 51.33\tiny${\textcolor{gray}{\pm0.06}}$ & \textbf{48.43}\tiny${\textcolor{gray}{\pm1.10}}$  & 35.47\tiny${\textcolor{gray}{\pm0.20}}$ & 52.17\tiny${\textcolor{gray}{\pm0.12}}$  & 27.71\tiny${\textcolor{gray}{\pm0.20}}$ & 33.81\tiny${\textcolor{gray}{\pm0.62}}$ & 83.98\tiny${\textcolor{gray}{\pm0.04}}$ & 91.30\tiny${\textcolor{gray}{\pm0.02}}$  & 49.62 & 56.43\\
		\texttt{CoT}  & 49.41\tiny${\textcolor{gray}{\pm1.02}}$ & 44.53\tiny${\textcolor{gray}{\pm5.27}}$ & 51.36\tiny${\textcolor{gray}{\pm1.53}}$ & 67.52\tiny${\textcolor{gray}{\pm1.32}}$  & 50.96\tiny${\textcolor{gray}{\pm0.32}}$ & \textbf{59.71}\tiny${\textcolor{gray}{\pm1.03}}$  & 82.01\tiny${\textcolor{gray}{\pm0.38}}$ & 90.04\tiny${\textcolor{gray}{\pm0.24}}$  & 58.44 & 65.45\\
		\texttt{\method{}}   & \textbf{52.46}\tiny${\textcolor{gray}{\pm0.43}}$ & 47.39\tiny${\textcolor{gray}{\pm1.59}}$ & \textbf{53.04}\tiny${\textcolor{gray}{\pm0.53}}$ & \textbf{69.11}\tiny${\textcolor{gray}{\pm0.48}}$  & \textbf{51.01}\tiny${\textcolor{gray}{\pm0.78}}$ & 59.18\tiny${\textcolor{gray}{\pm1.20}}$  & \textbf{84.33}\tiny${\textcolor{gray}{\pm0.02}}$ & \textbf{91.50}\tiny${\textcolor{gray}{\pm0.01}}$  & \textbf{60.21} & \textbf{66.80}\\

		\midrule
        \rowcolor[rgb]{ .980,  .980,  .980} 
        \multicolumn{11}{c}{\rule{0pt}{2ex}\textit{\fontsize{11pt}{11pt}\selectfont Qwen2.5-VL-7B-Instruct~\citep{qwen25vl}}}     \\  
		\midrule
		\texttt{Direct}  & 47.49\tiny${\textcolor{gray}{\pm0.22}}$ & 39.93\tiny${\textcolor{gray}{\pm4.33}}$  & 37.29\tiny${\textcolor{gray}{\pm0.80}}$ & 54.24\tiny${\textcolor{gray}{\pm0.95}}$  & 39.20\tiny${\textcolor{gray}{\pm0.29}}$ & 51.01\tiny${\textcolor{gray}{\pm0.55}}$  & 78.21\tiny${\textcolor{gray}{\pm0.27}}$ & 87.15\tiny${\textcolor{gray}{\pm0.20}}$  & 50.55 & 58.08\\
		\texttt{CoT}  & 57.86\tiny${\textcolor{gray}{\pm1.29}}$ & 43.00\tiny${\textcolor{gray}{\pm4.14}}$ & 55.84\tiny${\textcolor{gray}{\pm0.74}}$ & 71.29\tiny${\textcolor{gray}{\pm0.73}}$  & 44.11\tiny${\textcolor{gray}{\pm0.56}}$ & 57.96\tiny${\textcolor{gray}{\pm0.96}}$  & 83.65\tiny${\textcolor{gray}{\pm0.08}}$ & 90.63\tiny${\textcolor{gray}{\pm0.06}}$  & 60.37 & 65.72\\
		\texttt{\method{}}   &    \textbf{65.49}\tiny${\textcolor{gray}{\pm0.37}}$ & \textbf{52.81}\tiny${\textcolor{gray}{\pm0.68}}$ &    \textbf{63.78}\tiny${\textcolor{gray}{\pm0.19}}$ &    \textbf{77.86}\tiny${\textcolor{gray}{\pm0.17}}$  &    \textbf{66.94}\tiny${\textcolor{gray}{\pm0.21}}$ &    \textbf{79.13}\tiny${\textcolor{gray}{\pm0.23}}$  & \textbf{85.47}\tiny${\textcolor{gray}{\pm0.02}}$ & \textbf{91.67}\tiny${\textcolor{gray}{\pm0.01}}$  &    \textbf{70.42} &    \textbf{75.37}\\

        \midrule
        \rowcolor[rgb]{ .980,  .980,  .980} 
        \multicolumn{11}{c}{\rule{0pt}{2ex}\textit{\fontsize{11pt}{11pt}\selectfont Gemma3-12B-Instruct~\citep{gemma3}}}     \\  
		\midrule
		\texttt{Direct}  & 57.34\tiny${\textcolor{gray}{\pm0.44}}$ &    50.96\tiny${\textcolor{gray}{\pm3.74}}$  & 44.58\tiny${\textcolor{gray}{\pm0.55}}$ & 60.89\tiny${\textcolor{gray}{\pm0.59}}$  & 41.90\tiny${\textcolor{gray}{\pm0.40}}$ & 51.46\tiny${\textcolor{gray}{\pm0.87}}$ & 84.59\tiny${\textcolor{gray}{\pm0.07}}$ & 91.65\tiny${\textcolor{gray}{\pm0.04}}$  & 57.10 & 63.74\\
		\texttt{CoT}  & 57.69\tiny${\textcolor{gray}{\pm0.25}}$ & 49.92\tiny${\textcolor{gray}{\pm2.00}}$ & 60.56\tiny${\textcolor{gray}{\pm0.91}}$ & 75.09\tiny${\textcolor{gray}{\pm0.67}}$  & 60.32\tiny${\textcolor{gray}{\pm0.30}}$ & \textbf{72.24}\tiny${\textcolor{gray}{\pm0.83}}$  & 84.53\tiny${\textcolor{gray}{\pm0.33}}$ & 91.61\tiny${\textcolor{gray}{\pm0.19}}$  & 65.78 & 72.22\\
		\texttt{\method{}}   & \textbf{57.91}\tiny${\textcolor{gray}{\pm0.33}}$ & \textbf{52.54}\tiny${\textcolor{gray}{\pm0.77}}$ & \textbf{61.60}\tiny${\textcolor{gray}{\pm0.46}}$ & \textbf{75.81}\tiny${\textcolor{gray}{\pm0.51}}$  & \textbf{60.57}\tiny${\textcolor{gray}{\pm0.12}}$ & 70.97\tiny${\textcolor{gray}{\pm0.67}}$  & \textbf{85.40}\tiny${\textcolor{gray}{\pm0.13}}$ & \textbf{92.13}\tiny${\textcolor{gray}{\pm0.05}}$  & \textbf{66.37} & \textbf{72.86}\\

		\bottomrule
	\end{tabular}
	\end{adjustbox}
        \caption{Performance of MLLMs on multi-modal tasks in terms of Accuracy (\%) and Macro-F1. The \textbf{bold} indicates the best performance for the base LLM.}

    \label{mul results}
\vspace{-10pt}
\end{table*}

\subsection{Quantitative Comparison Results}
\label{sec:exp_quan_res}

% The quantitative results are presented in Table~\ref{mul results} for multi-modal tasks and Table~\ref{text results} for text-based tasks. The findings show that \method{} consistently outperforms both \texttt{Direct} and \texttt{CoT} by a clear margin across all tested LLMs, modalities, and benchmarks.

To ensure reliability and rule out the influence of generation randomness, we conduct all evaluations using three distinct random seeds. Consequently, we report the performance as the mean $\pm$ variance across these independent runs. The quantitative results are presented in Table~\ref{mul results} for multi-modal tasks and Table~\ref{text results} for text-based tasks. The findings show that \method{} consistently outperforms both \texttt{Direct} and \texttt{CoT} by a clear margin across all tested LLMs, modalities, and benchmarks.

\begin{table*}[t]
\vspace{-20pt}
\setlength{\tabcolsep}{7pt}
\centering
\renewcommand{\arraystretch}{1.2} 
\begin{adjustbox}{width=0.99\textwidth}
\begin{tabular}{lcccccccccc}
\toprule
\multirow{2}{*}{Model}  &   \multicolumn{2}{c}{\textbf{QuAC-1 (1,000)}} & \multicolumn{2}{c}{\textbf{QuAC-2 (2,000)}} & \multicolumn{2}{c}{\textbf{QuAC-3 (3,000)}} & \multicolumn{2}{c}{\textbf{QuAC (7,354)}} & \multicolumn{2}{c}{\textbf{SQuAD2.0 (11,873)}} \\
	\cmidrule{2-11}
           &  Acc  &  Macro-F1   & Acc & Macro-F1  &   Acc &  Macro-F1   &  Acc & Macro-F1  &  Acc & Macro-F1  \\

        \midrule
        \rowcolor[rgb]{ .980,  .980,  .980} 
        \multicolumn{11}{c}{\rule{0pt}{2ex}\textit{\fontsize{11pt}{11pt}\selectfont MiniCPM3-4B~\citep{minicpm}}}     \\  
		\midrule
		\texttt{Direct}  & 52.82\tiny${\textcolor{gray}{\pm0.62}}$ & 52.79\tiny${\textcolor{gray}{\pm0.40}}$  & 49.97\tiny${\textcolor{gray}{\pm0.34}}$ & 50.94\tiny${\textcolor{gray}{\pm0.23}}$  & 48.13\tiny${\textcolor{gray}{\pm0.36}}$ & 49.53\tiny${\textcolor{gray}{\pm0.23}}$  & 47.17\tiny${\textcolor{gray}{\pm0.19}}$ & 48.07\tiny${\textcolor{gray}{\pm0.05}}$  & 62.93\tiny${\textcolor{gray}{\pm0.15}}$ & \textbf{38.61\tiny${\textcolor{gray}{\pm0.08}}$}\\
		\texttt{CoT}  & 42.69\tiny${\textcolor{gray}{\pm0.94}}$ & 45.45\tiny${\textcolor{gray}{\pm0.71}}$ & 41.03\tiny${\textcolor{gray}{\pm0.33}}$ & 43.35\tiny${\textcolor{gray}{\pm0.73}}$  & 39.39\tiny${\textcolor{gray}{\pm0.27}}$ & 41.19\tiny${\textcolor{gray}{\pm0.74}}$  & 37.68\tiny${\textcolor{gray}{\pm0.18}}$ & 36.50\tiny${\textcolor{gray}{\pm0.20}}$  & \textbf{63.63\tiny${\textcolor{gray}{\pm0.11}}$} & 23.73\tiny${\textcolor{gray}{\pm0.04}}$\\
		\texttt{\method{}}   & \textbf{53.57\tiny${\textcolor{gray}{\pm0.41}}$} & \textbf{52.82\tiny${\textcolor{gray}{\pm0.43}}$} & \textbf{50.12\tiny${\textcolor{gray}{\pm0.26}}$} & \textbf{50.98\tiny${\textcolor{gray}{\pm0.25}}$}  & \textbf{48.53\tiny${\textcolor{gray}{\pm0.15}}$} & \textbf{49.75\tiny${\textcolor{gray}{\pm0.21}}$}  & \textbf{47.29\tiny${\textcolor{gray}{\pm0.27}}$} & \textbf{48.08\tiny${\textcolor{gray}{\pm0.30}}$}  & 62.74\tiny${\textcolor{gray}{\pm0.04}}$ & 36.83\tiny${\textcolor{gray}{\pm0.09}}$\\

        \midrule
        \rowcolor[rgb]{ .980,  .980,  .980} 
        \multicolumn{11}{c}{\rule{0pt}{2ex}\textit{\fontsize{11pt}{11pt}\selectfont Qwen2.5-7B-Instruct~\citep{qwen25}}}     \\  
		\midrule
		\texttt{Direct}  & 52.57\tiny${\textcolor{gray}{\pm0.39}}$ & 47.06\tiny${\textcolor{gray}{\pm0.20}}$  & 52.00\tiny${\textcolor{gray}{\pm0.20}}$ & 48.06\tiny${\textcolor{gray}{\pm0.10}}$  & 51.26\tiny${\textcolor{gray}{\pm0.22}}$ & 47.64\tiny${\textcolor{gray}{\pm0.15}}$  & 49.47\tiny${\textcolor{gray}{\pm0.07}}$ & 46.71\tiny${\textcolor{gray}{\pm0.15}}$  & 80.97\tiny${\textcolor{gray}{\pm0.13}}$ & \textbf{33.39\tiny${\textcolor{gray}{\pm0.06}}$}\\
		\texttt{CoT}  & 38.47\tiny${\textcolor{gray}{\pm1.55}}$ & 42.11\tiny${\textcolor{gray}{\pm0.90}}$ & 37.30\tiny${\textcolor{gray}{\pm1.16}}$ & 39.42\tiny${\textcolor{gray}{\pm0.76}}$  & 36.34\tiny${\textcolor{gray}{\pm0.68}}$ & 37.92\tiny${\textcolor{gray}{\pm0.22}}$  & 36.48\tiny${\textcolor{gray}{\pm0.29}}$ & 36.58\tiny${\textcolor{gray}{\pm0.09}}$  & 77.39\tiny${\textcolor{gray}{\pm0.14}}$ & 28.74\tiny${\textcolor{gray}{\pm0.06}}$\\
		\texttt{\method{}}   & \textbf{52.93\tiny${\textcolor{gray}{\pm0.25}}$} & \textbf{47.52\tiny${\textcolor{gray}{\pm0.21}}$} & \textbf{52.98\tiny${\textcolor{gray}{\pm0.10}}$} & \textbf{48.08\tiny${\textcolor{gray}{\pm0.07}}$}  & \textbf{52.01\tiny${\textcolor{gray}{\pm0.30}}$} & \textbf{47.77\tiny${\textcolor{gray}{\pm0.11}}$}  & \textbf{50.25\tiny${\textcolor{gray}{\pm0.21}}$} & \textbf{46.73\tiny${\textcolor{gray}{\pm0.15}}$}  & \textbf{81.04\tiny${\textcolor{gray}{\pm0.09}}$} & 29.63\tiny${\textcolor{gray}{\pm0.05}}$\\

        \midrule
        \rowcolor[rgb]{ .980,  .980,  .980} 
        \multicolumn{11}{c}{\rule{0pt}{2ex}\textit{\fontsize{11pt}{11pt}\selectfont Llama3.1-8B-Instruct~\citep{llama31}}}     \\  
		\midrule
		\texttt{Direct}  & 54.10\tiny${\textcolor{gray}{\pm0.02}}$ & 48.87\tiny${\textcolor{gray}{\pm0.14}}$  & 54.25\tiny${\textcolor{gray}{\pm0.08}}$ & 51.31\tiny${\textcolor{gray}{\pm0.09}}$  & 53.70\tiny${\textcolor{gray}{\pm0.04}}$ & 51.20\tiny${\textcolor{gray}{\pm0.13}}$  & 52.45\tiny${\textcolor{gray}{\pm0.07}}$ & 50.66\tiny${\textcolor{gray}{\pm0.15}}$  & 78.69\tiny${\textcolor{gray}{\pm0.02}}$ & 24.95\tiny${\textcolor{gray}{\pm0.04}}$\\
		\texttt{CoT}  & 54.90\tiny${\textcolor{gray}{\pm0.12}}$ & 46.09\tiny${\textcolor{gray}{\pm0.09}}$ & 51.75\tiny${\textcolor{gray}{\pm0.24}}$ & 38.21\tiny${\textcolor{gray}{\pm0.11}}$  & 49.13\tiny${\textcolor{gray}{\pm0.27}}$ & 34.47\tiny${\textcolor{gray}{\pm0.13}}$  & 46.95\tiny${\textcolor{gray}{\pm0.29}}$ & 28.74\tiny${\textcolor{gray}{\pm0.05}}$  & 58.19\tiny${\textcolor{gray}{\pm0.31}}$ & \textbf{28.86\tiny${\textcolor{gray}{\pm0.07}}$}\\
		\texttt{\method{}}   &    \textbf{59.40\tiny${\textcolor{gray}{\pm0.10}}$} &    \textbf{54.05\tiny${\textcolor{gray}{\pm0.04}}$} &    \textbf{58.05\tiny${\textcolor{gray}{\pm0.05}}$} &    \textbf{54.25\tiny${\textcolor{gray}{\pm0.08}}$}  &    \textbf{56.20\tiny${\textcolor{gray}{\pm0.09}}$} &    \textbf{53.45\tiny${\textcolor{gray}{\pm0.11}}$}  &    \textbf{53.51\tiny${\textcolor{gray}{\pm0.06}}$} &    \textbf{51.56\tiny${\textcolor{gray}{\pm0.09}}$}  & \textbf{79.83\tiny${\textcolor{gray}{\pm0.07}}$} & 25.72\tiny${\textcolor{gray}{\pm0.13}}$\\

		\bottomrule
	\end{tabular}
	\end{adjustbox}

        \caption{Performance of LLMs on text-based tasks in terms of Accuracy (\%) and Macro-F1. The \textbf{bold} indicates the best performance for the base LLM.}
        \vspace{-12pt}
    % \vspace{-0.8\baselineskip}
    \label{text results}
% \vspace{-5pt}
\end{table*}

The results demonstrate that \method{} consistently surpasses both \texttt{Direct} and \texttt{CoT} baselines across all tested LLMs (including Qwen2.5-VL and Gemma3) and benchmarks. Crucially, the low variance observed in \method{}'s performance indicates that anchoring the context via the attention sink not only improves accuracy but also enhances the stability of the model's reasoning process compared to standard prompting methods.

\textbf{Multi-modal Tasks.} 
On benchmarks designed to assess complex reasoning and object-level hallucination, \method{} shows substantial gains. For instance, on M3CoT, \method{} boosts the accuracy of \textit{Qwen2.5-VL-7B} by \textbf{+22.8}\% over \texttt{CoT}. On POPE, a benchmark specifically designed for hallucination detection, \method{} consistently improves both accuracy and Macro-F1 scores (\eg, \textbf{85.5}\% \textit{vs} 83.7\% accuracy and \textbf{91.8}\% \textit{vs} 90.6\% F1 score compared to \texttt{CoT}), confirming its effectiveness in mitigating the generation of non-existent objects.

\textbf{Text-based Tasks.} 
We use long-context conversational QA to specifically measure the mitigation of context forgetting. As the conversational context lengthens from QuAC-1 to QuAC, the performance of baseline methods degrades due to attention drift. In contrast, \method{} maintains a strong and often widening advantage. The most significant gains are seen with Llama3.1-8B, where on the longest-context QuAC setting (7,354 questions), \method{} achieves an accuracy of \textbf{56.2}\%, a \textbf{7.1}\% increase over \texttt{CoT}. On SQuAD2.0, which tests the LLM's ability to abstain from answering unanswerable questions, \method{} also yields consistent improvements.

\textbf{Analysis of CoT Performance.}
It is worth noting that CoT sometimes underperforms Direct Prompting, an observation consistent with existing studies. This performance degradation is highly task-dependent. In long-context understanding tasks, such as summarization and QA, CoT can even degrade performance~\citep{cotana3, cotana4}. Furthermore, while CoT excels in complex reasoning~\citep{cotana1} like M3CoT, in non-reasoning tasks, its extended chains exacerbate attention drift and raise the risk of error accumulation, where a single misstep derails the final answer~\citep{cotana2, cotana3}.

Taking together, these results validate that \method{} is a broadly effective and generalizable enhancement. It directly addresses both hallucination and context forgetting across different modalities, tasks, and model architectures without requiring any model training.

% The results show that \method{} outperformed the CoT baseline across the majority of the evaluated tasks and metrics. In multi-modal tasks, \method{} demonstrated improved capabilities in complex reasoning and hallucination suppression. Notably, on the M3CoT multi-modal complex reasoning benchmark, \method{} led to a 22.95 percentage point performance improvement for Qwen2.5-VL-7B-Instruct. In text-based long-context tasks, \method{} enabled Llama3.1-8B-Instruct to surpass the CoT baseline by 18.88 percentage points on the QuAC conversational question-answering benchmark. These results demonstrate the broad effectiveness and applicability of \method{} as a plug-and-play enhancement module across different modalities, tasks, and model architectures.

\vspace{-5pt}
\subsection{Qualitative Comparison Results}
\label{sec:exp_quali_res}

\begin{figure}[t]
    \centering
    \vspace{-18pt}
    \includegraphics[width=\linewidth]{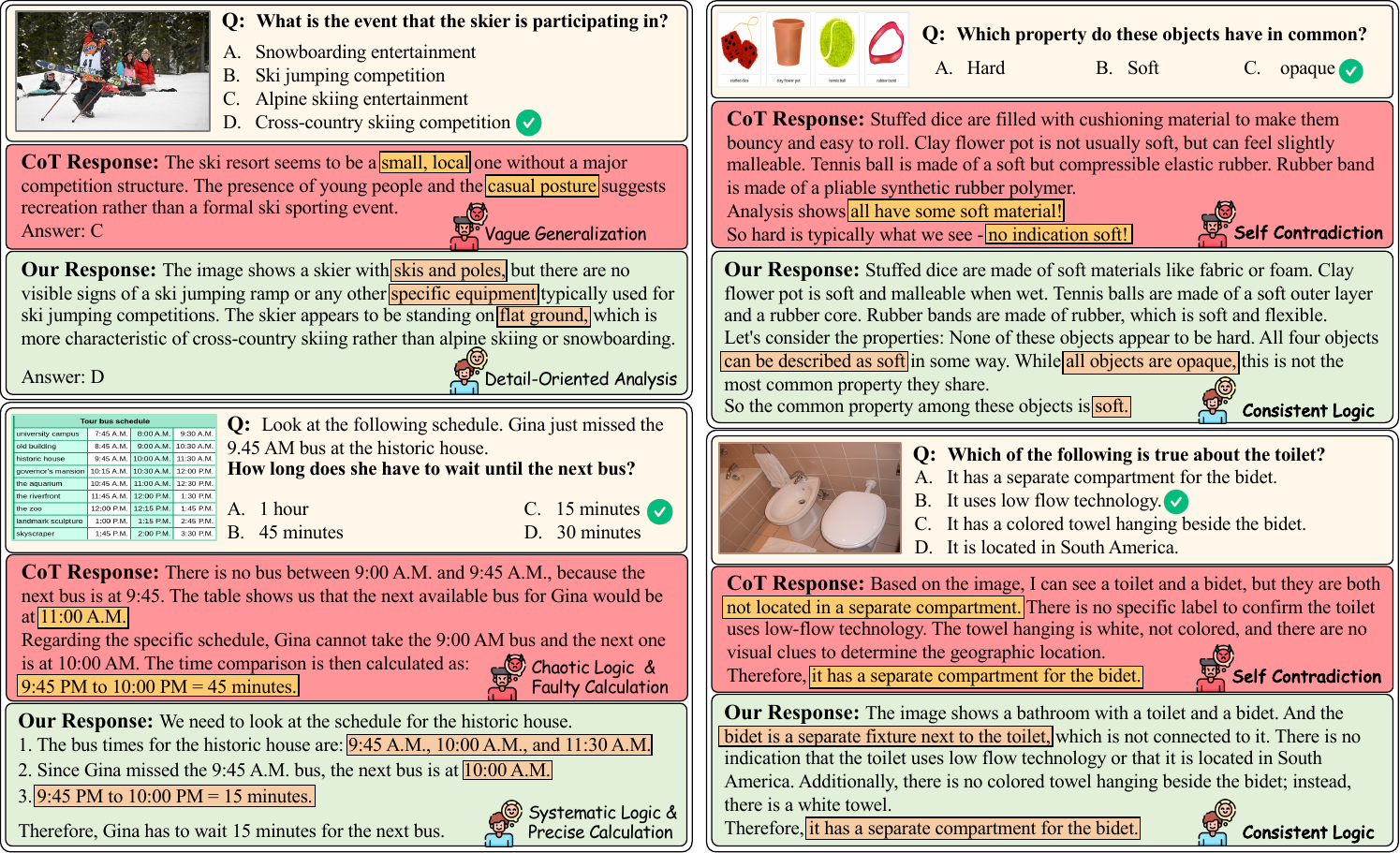}
    \vspace{-8pt}
    \caption{Qualitative comparison between \texttt{CoT} and \method{}.}
    \vspace{-10pt}
    \label{fig:case}
\end{figure}

\vspace{-5pt}
We present qualitative comparisons to illustrate how \method{} improves LLM's reasoning. In cases where \texttt{CoT} fails (Fig.~\ref{fig:case}, left), \method{} succeeds by grounding its analysis in specific contextual details (\eg, skis and poles) rather than ambiguous cues. This advantage also extends to failure cases where both methods produce an incorrect answer due to knowledge deficits (Fig.~\ref{fig:case}, right). Even then, \method{}'s reasoning chain remains internally consistent, whereas \texttt{CoT} exhibits context forgetting, where its final conclusion contradicts its own intermediate steps. This indicates that \method{} enhances the generation process by enforcing closer adherence to the initial context and maintaining logical integrity, even when LLM's underlying knowledge is flawed.

\vspace{-5pt}
\subsection{Analysis}\label{sec:analysis}
% To gain a deeper understanding of why \method{} is effective and to examine its underlying working mechanisms, we conducted a series of ablation studies, case studies, and quantitative analyses.

\subsubsection{Injection Strategies}
\label{sec:ablation}

\begin{wraptable}[9]{r}{0.33\textwidth}
    \renewcommand{\arraystretch}{1.2}
    \centering
    \small
    \captionsetup{font=small,width=0.32\textwidth}
    \setlength\tabcolsep{2pt}
    \vspace{-10pt}
% \vspace{-4pt}
    \resizebox{0.32\textwidth}{!}{
    \begin{tabular}{l c}
        \hline 
        \textbf{Method} & \textbf{Acc (\%)} \\ 
        \hline 
            \texttt{CoT} (Baseline) & 43.83 \\ 
            Inject Every Layer & 51.02 \\ 
            Inject Every 5 Layers & 60.04 \\ 
            Intermittent \& Increasing & 59.87 \\ 
            \rowcolor[rgb]{ .965, .965, .999 } 
            \textbf{Intermittent \& Decaying} & \textbf{60.69} \\ 
        \hline 
        \end{tabular} 
    }
    \vspace{-5pt}
    \caption{\!\!Performance \!of \!different \texttt{soft\! \!injection} \! variants\! (\S\ref{sec:soft}).}
    \label{tab:static_strategy}
\end{wraptable}

To study the effects of two main hyperparameters: the injection layers and the injection strength (\S\ref{sec:soft}), we analyze the performance of various \texttt{soft injection} using \textit{Qwen2.5-VL-7B-Instruct} on M3CoT. The results in Table~\ref{tab:static_strategy} highlight two critical findings: \textbf{i}) Injection Frequency. Intermittent injection (Inject Every 5 Layers) is more effective than continuous injection (Inject Every Layer), with a 9.0\% absolute accuracy improvement. This suggests that overly frequent intervention might disrupt LLM's native computational flow to some extent. \textbf{ii}) Injection Strength. A decaying schedule, which applies stronger injection in earlier layers, yields better performance (60.7\% \textit{vs} 59.9\%). This indicates that early contextual anchoring is more beneficial than later-stage interventions. In addition, the sensitivity to injection strength motivates us to develop an adaptive mechanism (\S\ref{sec:mean}).

\subsubsection{Hierarchical Attention Pattern}
\label{sec:spearman}

% \begin{wrapfigure}[12]{r}{0.44\textwidth}
%     \vspace{-17pt}
%     \centering
%     \small
%     \includegraphics[width=\linewidth]{figs/spearman.pdf}
%     \captionsetup{font=small,width=0.43\textwidth}
%     \vspace{-20pt}
%     \caption{xxxxxxxxxxxxxxx}
%     \label{fig:spearman}
% \end{wrapfigure}
To analyze the structural impact of our injection, we quantify its effect on LLM's attention pattern. Concretely, we compute the Spearman's rank correlation between $\langle$BOS$\rangle$'s attention weight vectors across all layers, before and after applying \method{}. The analysis reveals a near-perfect positive correlation ($\rho \!=\! 0.9985, p \!<\! 2.22 \!\times\! 10^{-17}$), as detailed in Appendix~\S\ref{sec:app_spearman}. This high correlation demonstrates that \method{} preserves the relative importance of attention across different layers. Rather than disrupting LLM's pre-trained attention hierarchy, \method{} enriches the information content of $\langle$BOS$\rangle$ while maintaining the integrity of the original computational flow.

% To verify that the injection operation of \method{} is non-disruptive, we quantified its impact on the model's pretrained attention pattern. We calculated the Spearman's rank correlation coefficient between the attention weight vectors for the initial token across all layers, before and after injection. As shown in Fig.~\ref{fig:spearman}, the correlation coefficient $\rho$ was $0.9985$ ($p < 2.22 \times 10^{-17}$), a value extremely close to 1. This demonstrates that \method{} preserves the stability of the model's inherent hierarchical attention pattern to a high degree during information injection. It does not alter the relative strength of attention weights across layers but instead smoothly guides the information flow upon this stable foundation, which is key to its safe and effective operation.

\subsubsection{Information Flow}
\label{ana:flow}
We further quantify the underlying working mechanism of \method{} as an information anchor by analyzing its vertical and information flows.

\begin{figure}[t]
    \centering
    \vspace{-24pt}
    \includegraphics[width=\linewidth]{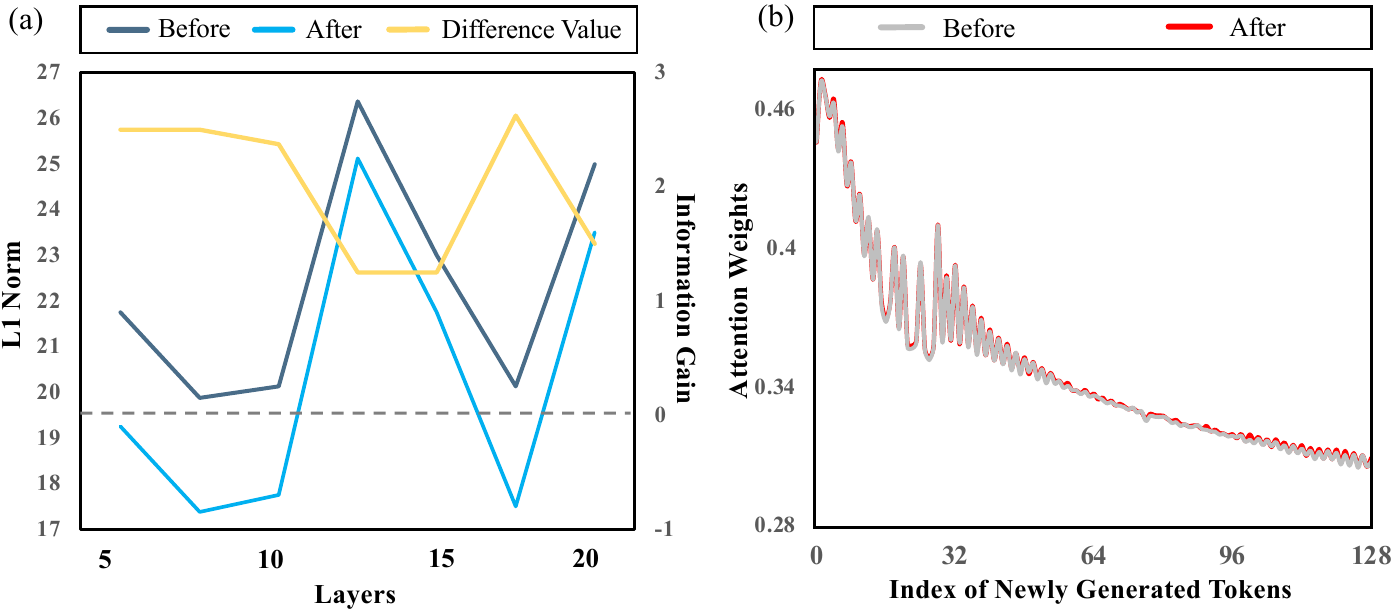}
    \vspace{-17pt}
    \caption{The illustration of information flow with and without \method{} (\S\ref{sec:mean}).}
    \label{fig:flowall}
    \vspace{-8pt}
\end{figure}

\textbf{Vertical Flow: Encoding via Value Vectors.}
The output of the attention mechanism is an aggregation of value vectors weighted by attention scores:
\begin{equation}
\bm{O}_t = \sum\nolimits_{j=1}^{t} \alpha_{t,j} \cdot \bm{V}_j,
\label{eq:attn_val}
\end{equation}
where $\alpha_{t,j}$ represents the attention weight and $\bm{V}_j$ denotes the value vector of the $j$-th token. Consequently, the information capacity of the $\langle$BOS$\rangle$ anchor can be quantified by calculating the L1 Norm of its value vector. As illustrated in Fig.~\ref{fig:flowall}(a), a significant difference value (yellow line) is observed between the original state (Before) and the injected state (After). This variation serves as a proxy for \textit{Information Gain}. The distinct divergence in L1 Norm across layers confirms that the anchor information is successfully absorbed and encoded into the $\langle$BOS$\rangle$ representation, effectively transforming it from a standard placeholder into a semantic anchor.

\textbf{Horizontal Flow: Sustained Attention Anchoring.}
To evaluate how this injected information is propagated during generation, we analyze the attention distribution of newly generated tokens. Specifically, we track the attention weights that subsequent tokens assign to the $\langle$BOS$\rangle$ anchor ($\alpha_{t, \text{BOS}}$). As shown in Fig.~\ref{fig:flowall}(b), the attention weights exhibit a natural decay but maintain a persistent magnitude, indicating that the generated sequence continuously attends to the $\langle$BOS$\rangle$ token. Crucially, the attention curves before and after injection (gray and red lines) overlap almost perfectly. This phenomenon demonstrates two key findings: \textbf{i}) \method{} successfully establishes a long-term information bridge, as the anchor remains a focal point throughout generation; and \textbf{ii}) The injection is non-destructive, preserving the intrinsic attention patterns of the LLMs.

\section{Related Work}

\textbf{Training-free Inference-time Enhancement for LLMs}.
Developing plug-and-play enhancement methods to improve the capabilities of LLMs has received widespread attention from the community. These approaches can be broadly categorized into three main streams. The first category is \textit{External Module Augmentation}~\citep{t1,wang2025recent,t2,t7,t17}. These methods couple an LLM with external components to address limitations such as outdated knowledge or the lack of specific skills, \eg, retrieval-augmented ~\citep{t1,t7,t10} and tool-augmented methods~\citep{t2,t4,t8,t17,cot}. The second category, \textit{Internal Reasoning Process Enhancement}~\citep{t6,t9,lei2024seeing,t20,t29,vitcot,weng2025we}, uses structured prompting strategies or adds reasoning steps to guide LLM, \eg, decomposing complex problems into simpler sub-tasks~\citep{t11,t17,cot,t23}, generating multiple candidate outputs with verification~\citep{t4,t12,t13,t29}, and using non-linear reasoning structures to organize complex inference paths~\citep{t20,t21,t22,cot}. The third category is \textit{Internal Representation Intervention}~\citep{t5,li2025red,t14,t16,t26,t28}, where methods directly modify the LLM's internal states during inference to guide its behavior, \eg, adding a targeted offset to LLM's hidden states~\citep{t5,t25,t26} and calibrating LLM's attention~\citep{t14,t15,t16,t27}. Despite impressive, these methods often require repetitive intervention, introducing considerable computational cost at each generation step.

Unlike methods that apply continuous modifications during the decoding loop, \method{} achieves training-free enhancement through a one-time intervention prior to inference. The injection of key information into the representation of $\langle$BOS$\rangle$ is a one-off operation performed before generation begins. Once this information anchor is established, the subsequent auto-regressive process is identical to the standard decoding procedure, requiring no extra computational steps.

\textbf{Attention Sink}
refers to the empirical observation that the initial token of a sequence (\ie, $\langle$BOS$\rangle$) consistently receives a high degree of attention during auto-regressive generation. This phenomenon is considered an emergent property of Transformer architectures~\citep{xiao2024efficientstreaminglanguagemodels,a2,a8}. Proposed causes include it being an artifact of the Softmax function within the attention mechanism~\citep{a14} or a consequence of the token's stable structural position in the sequence~\citep{t15,a12,a15}. Existing efforts use the attention sink in two main ways. One line of work uses the initial token's stability for tasks like KV cache management~\citep{a5} and key token protection~\citep{a6}. Another research direction focuses on enhancing LLM's performance by modifying or calibrating the attention mechanism to synergize with this structural anchor~\citep{t16,a8}.

This work explores a new application branch for attention sink: repurposing it as an active mechanism for context anchoring through information injection. We do not alter the attention weights themselves; instead, we modify the representation of $\langle$BOS$\rangle$ to form a high-information anchor that provides LLMs with high-integrity context. This shift from indirect attention calibration to direct information supply reveals a new direct to address the issues of hallucination and context forgetting.
\section{Conclusion}
This work mitigates two long-standing issues (\ie, hallucination and context forgetting) in decoder-only LLMs by exploiting the synergistic relationship between attention drift and attention sink. We propose \method{}, an efficient, plug-and-play method that transforms the passive attention sink into an active context anchor. Through an adaptive cross-attention mechanism, \method{} injects key contextual information into the first token's representation, thereby counteracting the information decay caused by attention drift during generation. Consistent performance improvements across diverse textual and multi-modal benchmarks validate its effectiveness and generalizability, and our analysis of its information delivery mechanism further substantiates the principles of \method{}.

% \method{} transforms the passive attention sink into an active context anchor, which is plug-and-play and adds negligible computational cost. Through an adaptive cross-attention mechanism,

% In conclusion, this study demonstrates that leveraging the intrinsic "Attention Sink" phenomenon is a training-free approach that enhances the reliability of LLMs. \method{} transforms the initial token into a dynamic information anchor.

% Our systematic exploration shows that:
% \textit{(i)} Direct "hard injection" fails due to structural disruption, while a "soft injection" approach, though beneficial, is limited by its sensitivity to manually tuned hyperparameters and its inability to adaptively select content.
% \textit{(ii)} The final adaptive mechanism, Attention Pin, resolves these limitations. It enhances complex reasoning, suppresses hallucinations, and mitigates context forgetting by dynamically retrieving information from unpooled sources, thereby maintaining the integrity of the injected content.
% \textit{(iii)} This enhancement is achieved with negligible computational overhead and exhibits broad generalizability across various model scales and modalities, an outcome attributed to its intervention at the fundamental level of the Transformer architecture.

% We expect that the empirical results and the "one-time pre-inference intervention" design presented in this work will encourage further research into efficient, training-free methods for enhancing the reliability and safety of LLMs.

\section*{Ethics Statement}
In this study, we propose \method{}, a training-free mechanism designed to enhance the reliability and reasoning capabilities of LLMs. We acknowledge that any technique that improves the fundamental capabilities of LLMs -- such as increasing their coherence or reducing factual errors -- carries a dual-use potential. While our work is intended for benevolent purposes, more capable and robust models could theoretically be exploited by malicious actors to generate more persuasive and harder-to-detect misinformation.

Our research is focused exclusively on the positive application of mitigating object hallucinations and improving long-context reasoning on established, public academic benchmarks. The datasets and tasks used in our experiments do not involve sensitive personal data, human subjects, or the generation of socially harmful content. Our methodology is transparent and designed to be analyzed and understood by the research community.

% We urge researchers and practitioners to consider the broader societal implications when developing and deploying enhanced AI systems. Our primary goal is to contribute to the development of safer, more robust, and more reliable AI. We encourage the responsible application of our findings and advocate for continued research into the safeguards and ethical considerations surrounding advanced AI technologies.

\section*{Reproducibility Statement} 
All experiments are conducted using publicly available, open-source LLMs and MLLMs. Specific model checkpoints are obtained from the official Hugging Face Hub. The detailed formulation and algorithmic implementation of \method{} are presented in \S\ref{sec:metho}. A comprehensive description of our experimental setup, including the specific datasets, evaluation metrics, and baselines used, is provided in \S\ref{sec:exp}. Any additional implementation details are further detailed in the Appendix. To facilitate full reproducibility, our complete source code, along with the scripts required to run all experiments and reproduce the results reported in this paper, is available at \href{https://github.com/67L1/SinkTrack}{GitHub}.

\bibliography{iclr2026_conference}
\bibliographystyle{iclr2026_conference}

\clearpage
\appendix
\centerline{\maketitle{\textbf{SUMMARY OF THE APPENDIX}}}
\renewcommand{\thetable}{S\arabic{table}}
\renewcommand{\thefigure}{S\arabic{figure}}
\setcounter{table}{0}
\setcounter{figure}{0}
\setcounter{algorithm}{0}

% This appendix contains additional details for the ICLR 2026 submission, titled \textit{``SinkTrack: Attention Sink based Context Anchoring for Large Language Models''}. It is organized as follows:

% \begin{itemize}[leftmargin=*]

%     \item \S\ref{sec:alg} offers the pseudo-code for \method{}.
%     \item \S\ref{sec:app_spearman} provides an analysis of \method{}'s impact on the hierarchical attention pattern.
%     \item \S\ref{sec:limitation} discusses the limitation of \method{}.
% \end{itemize}

\begin{itemize}[leftmargin=*]
\item \S\ref{sec:alg} details the pseudo-code implementation of \method{}.
% \item \S\ref{sec:exp_comprehensive} presents a comprehensive experimental evaluation, extending the main paper with additional benchmarks (LongBench, ScienceQA),and comparisons against stronger baselines (LongLM, InfLLM, etc.).
\item \S\ref{sec:analysis} provides in-depth analyses concerning the stability of the hierarchical attention pattern, resistance to attention drift, and computational overhead.
\item \S\ref{sec:limitation} discusses the limitations of the current approach.
\end{itemize}

\section{Pseudo-code for \method{}}\label{sec:alg}

To provide a clear description of \method{}, this section details the pseudo-code for it. As discussed in the main paper, the core idea is to create two parallel computational tracks within the attention module. \textit{Track $1$ (Injection Track)} uses cross-attention to inject external information into the initial token. \textit{Track $2$ (Original Track)} performs standard causal self-attention for the rest of the tokens to maintain the integrity of the pre-trained sequence structure. The complete process, including the conditional check for applying the injection, is described in Algo.~\ref{alg:hybrid_attention}.

\begin{algorithm}[h]
\renewcommand\thealgorithm{S1}
\caption{Pseudo-code for the Dual-Track Cross-Attention Mechanism.}
\label{alg:hybrid_attention}
\definecolor{codeblue}{rgb}{0.25,0.5,0.5}
\lstset{
  backgroundcolor=\color{white},
  basicstyle=\fontsize{7.2pt}{7.2pt}\ttfamily\selectfont,
  columns=fullflexible,
  breaklines=true,
  captionpos=b,
  escapeinside={(:}{:)},
  commentstyle=\fontsize{7.2pt}{7.2pt}\color{codeblue},
  keywordstyle=\fontsize{7.2pt}{7.2pt},
%  frame=tb,
}

\begin{lstlisting}[language=python]
"""
h_ori: hidden states of the original sequence ((:\color{codegreen}{$L_{ori} \times D_{h}$}):))
h_info: hidden states of the external information ((:\color{codegreen}{$L_{info} \times D_{h}$}):))
cfg: configuration for injection rules
"""
(:\color{codedefine}{\textbf{def}}:) (:\color{codefunc}{\textbf{Hybrid\_Attention}}:)(h_ori, h_info, cfg):
    #--- 1. Check if conditions for injection are met ---
    do_inject = (:\color{codecall}{\textbf{Check\_Conditions}}:)(cfg, h_info)

    (:\color{codedefine}{\textbf{if}}:) do_inject:
        #--- PATH A: HYBRID ATTENTION FOR INJECTION ---
        # 2a. Project inputs to Q, K, V spaces
        Q_ori, K_ori, V_ori = (:\color{codecall}{\textbf{Project\_QKV}}:)(h_ori)
        (:\color{codepro}{\textbf{K\_info}}:), (:\color{codepro}{\textbf{V\_info}}:)      = (:\color{codecall}{\textbf{Project\_KV}}:)(h_info)

        # 2b. Split the original Query for dual attention tracks
        (:\color{codedim}{\textbf{q\_first}}:), Q_rest = (:\color{codecall}{\textbf{Split\_Original\_Vector}}:)(Q_ori)

        # 3. Execute dual-track attention
        # Track 1: INJECTION via Cross-Attention
        o_first = (:\color{codecall}{\textbf{Cross\_Attention}}:)(Q=(:\color{codedim}{\textbf{q\_first}}:), K=(:\color{codepro}{\textbf{K\_info}}:), V=(:\color{codepro}{\textbf{V\_info}}:))

        # Track 2: CAUSAL SELF-ATTENTION
        m    = (:\color{codecall}{\textbf{Create\_Causal\_Mask}}:)(Q_rest)
        o_rest  = (:\color{codecall}{\textbf{Causal\_Self\_Attention}}:)(Q=Q_rest, K=K_ori, V=V_ori, mask=m)

        # 4. Merge outputs from both tracks
        o_combined = (:\color{codecall}{\textbf{Concatenate}}:)(o_first, o_rest)

    (:\color{codedefine}{\textbf{else}}:)
        #--- PATH B: STANDARD CAUSAL SELF-ATTENTION ---
        o_combined = (:\color{codecall}{\textbf{Standard\_Causal\_Attention}}:)(h_ori)
    
    # 5. Apply final output projection
    h_out = (:\color{codecall}{\textbf{Output\_Projection}}:)(o_combined)
    
    (:\color{codedefine}{\textbf{return}}:) h_out

\end{lstlisting}
\end{algorithm}

\section{In-depth Analysis: Mechanism and Efficiency}\label{sec:analysis}
This section provides a deeper analysis of the underlying mechanism of \method{} and its computational implications.

\subsection{Stability of Hierarchical Attention Pattern}
\label{sec:app_spearman}

\begin{figure}[t]
    \centering
    \small
    \includegraphics[width=0.5\linewidth]{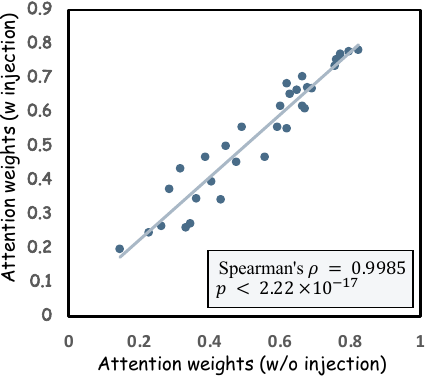}
    \captionsetup{font=small,width=1.0\textwidth}
    % \vspace{-20pt}
    \caption{Comparison between the $\langle$BOS$\rangle$'s attention weight vectors  before and after applying \method{}.}
    \label{fig:spearman}
\end{figure}

To analyze the structural impact of our injection, we quantify its effect on the LLM's hierarchical attention pattern. Concretely, we compute the Spearman's rank correlation between the $\langle$BOS$\rangle$'s attention weight vectors across all layers, before and after applying \method{}. As shown in Fig.~\ref{fig:spearman}, the analysis reveals a near-perfect positive correlation ($\rho=0.9985, p < 2.22 \times 10^{-17}$).

This high correlation demonstrates that \method{} preserves the relative importance of attention across different layers. Rather than disrupting the LLM's pre-trained attention hierarchy pattern, \method{} enriches the information content of the initial token while maintaining the integrity of the original computational flow. This finding provides a quantitative explanation for why \method{} can enhance performance without causing the model collapse observed in \texttt{hard injection}.

\subsection{Verification of \method{}'s Stability}
To further verify that \method{} remains effective over long generation sequences, we conduct a ``Drift Test'' on Llama3.1-8B. We feed it an input context of 8k+ tokens and track the attention weights allocated to the $\langle$BOS$\rangle$ token by subsequently generated tokens (steps 1 to 1024).

As presented in Table~\ref{tab:drift_test}, the attention weight to $\langle$BOS$\rangle$ remains dominant throughout the generation process. For instance, at generated token index 1024 (sequence position 9106), the attention to $\langle$BOS$\rangle$ is 0.582, which is approximately $14\times$ higher than the maximum attention paid to any other token. This confirms that \method{} creates a robust attention sink that is immune to attention drift, ensuring long-term context retention.

\begin{table}[h]
\centering
\small
\caption{SinkTrack Drift Test on Llama3.1-8B. The attention to $\langle$BOS$\rangle$ consistently dominates other tokens even after generating 1024 new tokens.}
\label{tab:drift_test}
\begin{tabular}{ccccc}
\toprule
\textbf{Gen. Idx} & \textbf{Seq. Pos} & \textbf{Attn to $\langle$BOS$\rangle$} & \textbf{Max Attn (Others)} & \textbf{Ratio} \\
\midrule
1 & 8083 & 0.561 & 0.057 & 9.87x \\
64 & 8146 & 0.570 & 0.048 & 11.97x \\
128 & 8210 & 0.612 & 0.054 & 11.25x \\
256 & 8338 & 0.560 & 0.050 & 11.24x \\
512 & 8594 & 0.558 & 0.037 & 15.17x \\
1024 & 9106 & 0.582 & 0.042 & 13.96x \\
\bottomrule
\end{tabular}
\end{table}

\subsection{Computational Overhead}
We measure the prefill stage latency on an NVIDIA GeForce RTX 4090 to evaluate the computational cost of \method{}. The results, averaged over 500 samples, are presented in Table~\ref{tab:overhead}. Specifically, Llama3.1-8B shows a latency increase from 35.90 ms \textit{(w/o Injection)} to 36.66 ms \textit{(w/ Injection)}, representing a marginal overhead of only +0.76 ms. Similarly, Qwen2.5-VL-7B shows an increase of +3.13 ms. These findings confirm that \method{} introduces negligible latency, as the core intervention is a one-off operation performed prior to the auto-regressive generation.

\begin{table}[h]
\centering
\small
\caption{Prefill stage latency measurement on NVIDIA GeForce RTX 4090 (averaged over 500 samples). \method{} introduces negligible computational overhead.}
\label{tab:overhead}
\begin{tabular}{lccc}
\toprule
\textbf{Model} & \textbf{w/o Injection} & \textbf{w/ Injection} & \textbf{Overhead} \\
\midrule
Llama3.1-8B & 35.90 ms & 36.66 ms & +0.76 ms \\
Qwen2.5-VL-7B & 107.13 ms & 110.26 ms & +3.13 ms \\
\bottomrule
\end{tabular}
\end{table}

\section{Limitation}\label{sec:limitation}
This work proposes and validates \method{} representing a solid step towards mitigating hallucination and context forgetting in LLMs. However, it is important to recognize several inherent limitations in our current approach that open avenues for future research. Specifically, \method{}'s core design anchors all contextual information to the representation of a single initial token. While our adaptive mechanism optimizes information retrieval, this setup introduces a potential information bottleneck, as a single fixed-dimension vector has a finite capacity for representation. This may not fully capture the nuanced informational needs of tasks requiring extremely long or dense contexts. For example, while effective, the diminishing performance gains we observe on the long-dialogue QuAC may be an early indicator of this bottleneck. This limitation suggests a need to explore future paradigms beyond a single anchor, such as distributed or dynamically positioned anchoring mechanisms.

% Moreover, the effectiveness of \method{} is inherently tied to the quality and nature of the user-provided context from which it draws information. Our current implementation directly injects features derived from this raw context. While this approach ensures that the model is grounded in the provided source, it does not guarantee that this source contains the most effective information for solving a given task.
% For tasks demanding high-level or abstract reasoning, injecting raw features is akin to providing a collection of unprocessed evidence. A more effective strategy might involve injecting a pre-computed summary, a logical plan, or a structured knowledge representation. For instance, when solving a complex multi-modal math problem, supplementing the raw image and question with an injected "step-by-step solving plan" (potentially generated by another LLM) could be significantly more beneficial than injecting the raw inputs alone. Exploring how \method{}'s efficient injection mechanism can be coupled with more advanced techniques for context abstraction and refinement remains a significant and promising direction for future work.

\end{document}